\RequirePackage{fix-cm}
\documentclass[journal]{IEEEtran}         % twocolumn
\ifCLASSINFOpdf
\else
\fi
\usepackage{graphicx}
\usepackage{tabularx}
\usepackage{subfigure}  % support sub-figure
\usepackage{adjustbox}
\usepackage{mathrsfs}
\usepackage{amsmath}
\usepackage{multirow}
\usepackage{mathrsfs}
\usepackage{algpseudocode}
\usepackage{amsmath} 
\usepackage{amsfonts}
\usepackage{amssymb}
\usepackage{bbm}
\usepackage{algorithmicx,algorithm}
\usepackage[misc]{ifsym}
\usepackage[square,sort,comma,numbers]{natbib}
\usepackage{caption}
\usepackage{multirow}
\usepackage[normalem]{ulem}
\useunder{\uline}{\ul}{}
\usepackage{latexsym}
\def\x{{\mathbf x}}
\begin{document}
\title{Fully Unsupervised Person Re-identification via Selective Contrastive Learning}
\author{Bo Pang,
        Deming~Zhai,~\IEEEmembership{Member,~IEEE,}
        Junjun Jiang,~\IEEEmembership{Member,~IEEE,}
        and~Xianming~Liu,~\IEEEmembership{Member,~IEEE}% <-this % stops a space
\thanks{B. Pang, D. Zhai, J. Jiang and X. Liu are with the School of Computer Science and Technology, Harbin Institute of Technology, Harbin 150001, China, and also with the Peng Cheng Laboratory, Shenzhen 518052, China  (cspb@hit.edu.cn; zhaideming@hit.edu.cn; jiangjunjun@hit.edu.cn; csxm@hit.edu.cn).}% <-this % stops a space
% \thanks{J. Doe and J. Doe are with Anonymous University.}% <-this % stops a space
% \thanks{Manuscript received April 19, 2005; revised August 26, 2015.}
}
% \institute{
% Bo Pang 
% \email{19b903944@stu.hit.edu.cn}
% \and
% Deming Zhai\\
% \email{zhaideming@hit.edu.cn}
% \and
% Xianming Liu\\
% \email{csxm@hit.edu.cn}
% \and
% Junjun Jiang\\
% \email{jiangjunjun@hit.edu.cn}
%  B. Pang, D. Zhai, J. Jiang and X. Liu are with the School of Computer Science and Technology, Harbin Institute of Technology, Harbin 150001, China, and also with the Peng Cheng Laboratory, 
% }

\maketitle
\markboth{Journal of \LaTeX\ Class Files,~Vol.~14, No.~8, August~2015}%
{Shell \MakeLowercase{\textit{et al.}}: Bare Demo of IEEEtran.cls for IEEE Journals}

\begin{abstract}
% Person re-identification has attracted lots of attention in recent years. This paper proposes a joint global and local learning framework for tackling the unsupervised person re-identification. Existing unsupervised studies usually adapt CNN to extract feature maps on the person images and use GAP to get the global features. These approaches attempt to solve unsupervised person re-identification by  similarity learning with the global features. However, the global features obtained by GAP will loss information of the images while the local features could keep more details of the person images. Despite the local features are better representation for the person images, these local parts are independent and have no connection, which makes the unsupervised learning process hardly converge to be satisfied. But the global features could guide the local features with providing a roughly feasible convergence direction for each local part. In such case, we set a mixture memory bank to connect the global features and the local features, where the local features will make up the low representation ability for the global features  and the global features will guide and unify each part of the local features. We test our approach on two image-based and two video-based datasets and the performance has achieved state-of-the-art.

Person re-identification (ReID) aims at searching the same identity person among images captured by various cameras. Existing fully-supervised person ReID methods usually suffer from poor generalization capability caused by domain gaps. Unsupervised person ReID attracts a lot of attention recently, due to it works without intensive manual annotation and thus shows great potential of adapting to new conditions. Representation learning plays a critical role in unsupervised person ReID. In this work, we propose a novel selective contrastive learning framework for fully unsupervised feature learning. Specifically, different from traditional contrastive learning strategies, we propose to use multiple positives and adaptively selected negatives for defining the contrastive loss, enabling to learn a feature embedding model with stronger identity discriminative representation. Moreover, we propose to jointly leverage global and local features to construct three dynamic memory banks, among which the global and local 
ones are used for pairwise similarity computation and the mixture memory bank are used 
for contrastive loss definition. Experimental results demonstrate the superiority of our method in unsupervised person ReID compared with the state-of-the-arts.

\end{abstract}

\IEEEpeerreviewmaketitle
\section{Introduction}

\IEEEPARstart{P}{erson} re-identification (ReID), also referred to as person retrieval, aims at searching the same identity person among images captured by various cameras at different time and locations. 
Thanks to the rapid development of convolutional neural networks (CNN), in recent years, the performance of person ReID is improved remarkably by using discriminative features from labeled person images. However, the success of such systems relies on a large amount of labeled data, which is often prohibitively expensive to acquire. As a result, a large research effort is currently focused on unsupervised systems without leveraging intensive manual supervision, which attracts a lot of attention due to the great potential of adapting to new conditions. 
	
% In this paper, we concentrate on self-supervised visual representation learning.	
% 	It attracts a lot of attention due to its great potential application value. Many works attempt to solve this problem in a supervised way \citepp{luo2019bag,sun2018beyond,li2014deepreid,varior2016gated} by annotating
% 	each person images and  adapt the convolutional neural network to learn discriminative features from these images. However, such supervised manners  are much too expensive. Therefore, how to utilize unlabeled data to solve  person re-identification  has become an important research area. In this paper, we are going to  tackle the problem under the unsupervised setting.

This effort includes recent advances on transfer learning and unsupervised learning. Among them, cross-domain transfer learning, also called domain adaptation, offers an effective manner to reduce the labelling cost. The basic idea is to learn an attribute-semantic and identity discriminative representation from a labeled source dataset, which is then transferred to the target domain for person ReID \cite{wang2018transferable,deng2018image,liu2019adaptive,ge2020selfpaced}. However, the performance of this approach 
largely relies on the assumption that there is sufficient knowledge overlap between the source and target domains, which is not always valid however. If the source domain exhibits significantly different characteristics with the target domain, the ReID performance would degrade heavily.  Moreover, this kind of methods require a large amount of annotated source data, which are thus not purely unsupervised. 
% 	To surmount the difficulties on unsupervised person ReID, some methods \citep{yu2017cross,zhong2019invariance,lv2018unsupervised} adapt transfer learning, leveraging other dataset's labeled information to conquer the challenge with unlabeled data on the target dataset. Such methods provide novel ideas for unsupervised person ReID but still used labeled data.  

The fully unsupervised learning based approach receives more and more attention recently, since it works without leveraging any labeled data and thus shows better potential to deploy person ReID for real-world applications. The basic idea of this line is to alternate between predicting pseudo labels by clustering or classification and training the network with generated pseudo classes. For instance, Lin \textit{et al.} \citep{lin2019bottom} proposed a bottom-up clustering framework that iteratively trains a network based on the pseudo labels generated by unsupervised clustering. Wang \textit{et al.} \citep{wang2020unsupervised} formulated unsupervised person ReID as a multi-label classification task to progressively seek true labels. Lin \textit{et al.} \citep{lin2020unsupervised} proposed a classification network with softened labels to eliminate 
errors incurred from hard quantization in clustering. However, the performance of these methods relies on the accuracy of label prediction, which is a non-trivial task under unsupervised setting.

Instead of explicit label prediction, in this paper, we concentrate on contrastive self-supervised visual representation learning, taking advantage of the principle that a good feature representation model should map images of the same person closer to each other, while push images of different identities apart away. Specifically, we propose a novel selective contrastive learning framework with 
dynamic memory banks, which is specially tailored for the task of unsupervised person ReID. Although there is also some work attempts to tackle person ReID based on contrast learning such as \cite{ge2020selfpaced}, which is upon the domain adaption paradigm as thus requires a labeled source domain. In contrast, the proposed method is purely unsupervised.

The technical contributions of this work are three-fold: 
\begin{itemize}
		\item Considering that a person may be captured by various cameras, \textit{i.e.}, each person may have multiple images in the training set, one key technical novelty is that we choose multiple positives for each anchor, as opposed to SimCLR \citep{chen2020simple}, MoCo \citep{he2020momentum} and \cite{ge2020selfpaced} that use only a single positive to define the contrastive loss. Moreover, different from the conventional contrastive learning strategies that take all samples except the positive as the negatives \citep{chen2020simple,he2020momentum,ge2020selfpaced}, we propose to select samples that are plausibly similar to the anchor as the negatives, so as to improve the discrimination ability of representation learning. More specifically, 
	using the defined distance metric, we rank the similarity order between the anchor and all training samples, according to which we divide the training set into three subsets: \textit{similar set}, 
\textit{borderline set}, and \textit{dissimilar set}. By taking samples in \textit{similar set} as the positives and samples in \textit{borderline set} as the negatives, we define the contrastive loss to encourage the feature embedding function to produce closely aligned representations to all images of the same identity.

\item  The other contribution of this work is that we propose three dynamic dictionaries which jointly leverage global and local discriminative information for unsupervised representation learning.  The global feature is widely used in existing unsupervised ReID, such as \citep{lin2019bottom,wang2020unsupervised}. However, it suffers from discriminative information loss in some cases, leading to images of different persons may have similar feature representations. On the other hand, the part-level features offer fine-grained discriminative information for pedestrian image description \citep{sun2018beyond}. However, compared with the global feature, the local features bring much more search freedom, making the optimization of feature representation learning hard to converge. Moreover, learning discriminative local features requires that parts should be accurately located. It can be done either by external assistance from human pose estimation \citep{poseestimate} or well-designed partition strategies \citep{Zhao_2017_ICCV,sun2018beyond}, which are expensive and hinder the generalization in practical applications. 
Considering their respective limitations, we propose to jointly leverage the global feature and the local features for defining distance metric and contrastive loss.

\item By combining the above contributions, we propose an effective unsupervised person ReID algorithm. Our scheme achieves encouraging performance with respect to rank-1 and mAP so far on public Market-1501, DukeMTMC-reID, DukeMTMC-VideoReID and MARS. For instance, we achieve rank-1 accuracy of 82.2\%, significantly outperforming the latest unsupervised person ReID methods SNR \citep{2020Style}, SSLR \citep{lin2020unsupervised}, MMCL \citep{wang2020unsupervised} and TSSL \citep{2020Tracklet} by 15.5\%, 10.5\%, 15.6\% and 11\%, respectively.
		
		\end{itemize}

The paper is organized as follows: Section 2 overviews some related works. Section 3 introduces the proposed scheme, including the representation learning framework, positives and negatives sampling strategy, the defined contrastive loss and the optimization strategy, and the memory bank update strategy.  Section 4 provides the experimental results and ablation study.  We conclude this paper in Section 5.

% 	\begin{figure}[t] 
% 		\centering  

% 		\includegraphics[width=1\columnwidth]{./a3.png}
	
% 		\caption{An example illustration (Optimal features indicate the results of any well-trained model). Part-based methods are proved useful for person ReID. However, under the fully unsupervised settings, local features will bring much more search freedom. With global features guided, this problem could be avoided. }
% 		\label{a1}
% 	\end{figure}
	
% 	\begin{figure}[t] 
% 		\centering  

% 		\includegraphics[width=0.7\columnwidth]{./a1.png}

% 		\caption{An example illustration. The first row shows that global average pooling (GAP) would lead two different feature maps become the same; while as shown in the second row, the local average pooling (LAP) can avoid this problem. }
% 		\label{a1}
% 	\end{figure}
	
\section{Related Work}
% In this section, we provide an overview about person ReID.

\subsection{Unsupervised Domain Adaptation Person ReID}

Transfer learning is a common  strategy for addressing unsupervised person ReID. 
These domain adaption methods \citep{wang2018transferable,deng2018image,liu2019adaptive,zhong2018generalizing,2020Style} attempt to tackle the unsupervised person ReID problem on the target unlabeled dataset by leveraging other dataset's labeled information. TJ-AIDL model \citep{wang2018transferable} aims at learning an attribute-semantic and identity-discriminative feature representation space which is transferrable to the any unlabelled target dataset. HHL \citep{zhong2018generalizing} aims to improve the generalization ability of re-ID models on the target testing set with enforcing two properties, camera invariance and domain  connectednes, simultaneously. Thanks to the development of the Generative Adversarial Network (GAN), this type of style transfer network is used for cross-domain transfer learning for unsupervised person ReID. SPGAN \citep{deng2018image} generates transferred images from the source labeled dataset and then do the supervised learning with two constraints which are self-similarity of an image before and after translation and domain-dissimilarity of a translated source image and a target image. ATNet \citep{liu2019adaptive} decomposes the complicated cross-domain transfer into a set of factor-wise sub-transfers, each of which concentrates on style transfer with respect to a certain imaging factor, e.g., illumination, resolution and camera view etc. Considering poor generalization capability caused by domain gaps with existing methods, Baseline-SNR \citep{2020Style} filter out identity-irrelevant interference and learn domain-invariant person representations. In \cite{ge2020selfpaced}, a hybrid memory is proposed to encode all available information from both source and target domains for feature learning, which achieves the best unsupervised person ReID performance so far.
Although achieving promising performance, these methods require an annotated source dataset. In contrast, our work focuses on purely unsupervised person ReID, which only relies on the unlabeled target dataset.

\subsection{One Shot Person ReID}
Methods based on one-shot learning for person ReID attempt to solve the problem with condition where each identity has only one labeled example and many unlabeled examples. EUG \citep{2018Exploit} and ProLearn \citep{2019Progressive} propose to gradually and steadily improve the discriminative capability of the CNN via stepwise learning. Especially, for video-based person re-identification, RACE \citep{ye2018robust} firstly adopt anchor sequences to formulate an anchor graph. And then for accurately estimate labels from unlabeled sequences with noisy frames, robust anchor embedding is introduced based on the regularized affine hull. These methods solved the cost of annotation in a degree compared with the supervised methods and the domain adaptation based methods. But still, the labeled information is needed. 

\subsection{Fully Unsupervised Person ReID}
For fully unsupervised person ReID, traditional methods \citep{liao2015person,zheng2015scalable} utilize  hand-craft features, which are hardly designed to be discriminative by hand. Recently, the cluster based methods and the mutli-label based methods estimate pseudo labels to train the neural network. BUC \citep{lin2019bottom} jointly optimize a convolutional neural network (CNN) and the relationship among the individual samples with bottom-up clustering procedure. TSSL \citep{2020Tracklet} thought these bottom-up clustering methods merely utilise suboptimal global clustering. They design a comprehensive unsupervised learning objective that accounts for tracklet frame coherence, tracklet neighbourhood compactness, and tracklet cluster structure in a unified formulation, which is capable of capitalising directly from abundant unlabelled tracklet data. Wang \textit{et al.} formulate the problem as a multi-label classification task to progressively seek true labels and adopt the memory-based multi-label classification loss (MMCL) to boost the ReID model training efficiency. SSLR \citep{lin2020unsupervised} proposed a similarity learning framework with softened labels to relief the hard quantization loss in clustering.  However, these methods rely on the accuracy of the pseudo labels. Our proposed framework will adopt contrastive self-supervised visual representation learning, mapping the person images with same identity close to each other while push the person with different identity apart away. 

\begin{figure*}[t]

\centering
\includegraphics[width=1.0\linewidth]{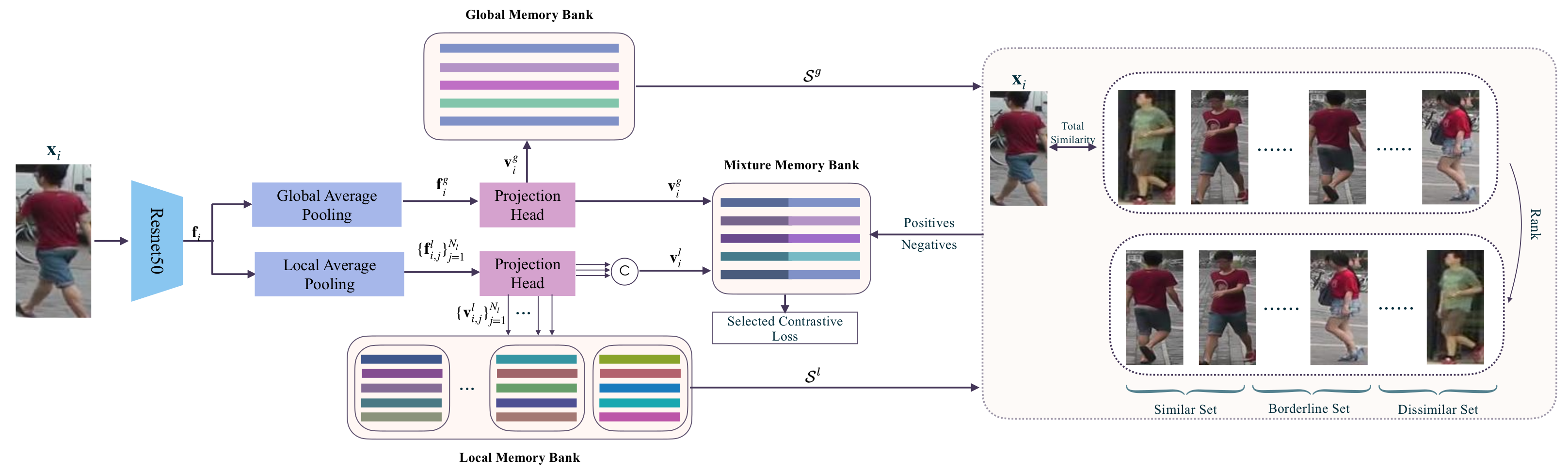}% 1\linewidth
\caption{The overall framework of our proposed unsupervised person ReID.}
\label{b1}

\end{figure*}

\subsection{Unsupervised Representation Learning}

Unsupervised representation learning utilize unlabeled data to learn an effective embedding space for downstream tasks like image classification and etc. Gidaris \textit{et al.} \citep{2018spUnsupervised} propose to learn image features by training ConvNets to recognize the 2d rotation that is applied to the image that it gets as input. Recently, contrastive learning has attracted a lot of attention. Such methods aim at mapping the representation  close to the positives and away from the negatives. MoCo \citep{he2020momentum} build a large and consistent dictionary on-the-fly that facilitates contrastive unsupervised learning. SimCLR \citep{chen2020simple} propose  a simple framework for contrastive learning of visual representations. Both the Moco and the SimCLR  use only one positive to conduct contrastive loss. In our paper, we propose a new contrastive learning framework with selective positives and negatives and three dynamic dictionaries will be conducted to help the training process.

% Unsupervised representation learning utilize unlabeled data to learn an effective embedding space for downstream tasks like image classification and etc. Gidaris \textit{et al.} \citep{2018spUnsupervised} propose to learn image features by training ConvNets to recognize the 2d rotation that is applied to the image that it gets as input. Wu \textit{et al.} \citep{2018Unsupervised} design a non-parametric classifier at the instance-level. Recently, contrastive learning has attracted a lot of attention. Such methods aim at mapping the representation  close to the positives and away from the negatives. MoCo \citep{he2020momentum} build a large and consistent dictionary on-the-fly that facilitates contrastive unsupervised learning. SimCLR \citep{chen2020simple} propose  a simple framework for contrastive learning of visual representations. Both the Moco and the SimCLR  use only one positive to conduct contrastive loss. In our paper, we propose a new contrastive learning framework with selective positives and negatives and three dynamic dictionaries will be conducted to help the training process.

\section{Methodology}

Given a training set $\mathcal{X} = \{\x_1, \x_2, ..., \x_{N}\}$, the goal of unsupervised person ReID is to learn a model $\mathcal{F}(\boldsymbol{\theta};\x_i)$ for visual feature representation without using any manual annotation, where parameters related to $\mathcal{F}$ are denoted as $\boldsymbol{\theta}$. The learned representation model is then applied on the query image $\x^q$ and the gallery set $\hat{\mathcal{X}} = \{\hat{\x}_1, \hat{\x}_2, ..., \hat{\x}_{M}\}$, so as to derive the query result by ranking distance between features of the query and all gallery images. It is clear to see that, unsupervised representation learning plays a central role in unsupervised person ReID. 

The feature representation learning is challenging in unsupervised setting, wherein the critical problem is how to learn to discriminate between individual images, without any notion of semantic categories. A good feature representation model should map images of the same person closer to each other, while push images of different identities apart away. In this work, inspired by recent progress of self-supervised learning, we address the unsupervised feature learning problem from the perspective of contrastive learning.

	\subsection{Representation Learning Framework}
	
	As illustrated in Fig. \ref{b1}, our representation learning framework consists of the following components:
	\begin{itemize}
\item \textit{A neural network base encoder} $\mathbf{E}(\cdot)$, which is leveraged to extract feature vector for a given person image $\x_i$. It allows various choices of the network architecture without any constraints. Here we adopt ResNet-50 without the last classification layer as the backbone, whose parameters are pre-trained on ImageNet, to obtain feature map $\mathbf{f}_i = \mathbf{E}(\x_i) \in \mathbb{R}^{2048}$.

\item \textit{Pooling operator} $\mathbb{AP}(\cdot)$, for which we consider both global and local average pooling to obtain global and local features. The global feature $\mathbf{f}^g_i$ is obtained by average pooling (AP) of feature map $\mathbf{f}_i$ of the whole image: 
		\begin{equation}
	\mathbf{f}^g_i = \mathbb{AP}(\mathbf{f}_i)
	\end{equation}
	 However, the global feature suffers from discriminative information loss in some cases, leading to images of different identities may have similar feature representation. We further consider the part-level features, which offer fine-grained discriminative information for pedestrian image description \citep{sun2018beyond}.  We obtain part-level feature maps by equally partitioning the global feature map $\mathbf{f}_i$ into $N_l=8$ horizontal stripes. The local features are then obtained by 
	applying average pooling on each part:
	\begin{equation}
	\mathbf{f}^l_{i,j} = \mathbb{AP}(\mathbf{f}_{i,j}), j=1,\cdots,N_l
	\end{equation}

	\item \textit{A small projection neural network} $\mathbf{P}(\cdot)$, which is a learnable nonlinear operation that transforms image features to the latent space where the contrastive learning is conducted. $\mathbf{P}(\cdot)$ is defined with fully-connected (FC) layer, batch normalization (BN) layer and L2-normalization (L2-Norm) layer. The usage of $\mathbf{P}(\cdot)$ is demonstrated to be beneficial to define the contrastive loss \citep{chen2020simple}.

	\item \textit{Global and local memory bank} $\mathcal{M}^g$ and $\mathcal{M}^l$, which are  leveraged to store global and local features for pairwise similarity computation. The keys in global memory bank $\mathcal{M}^g$ are defined as:
	\begin{equation}
	\label{global}
	\mathbf{v}^g_i = \mathbf{P}\left(\mathbb{AP}(\mathbf{f}_i)\right), i=1,\cdots,N
	\end{equation}
	And the keys in local memory bank $\mathcal{M}^l$ are defined as:
	\begin{equation}
	\label{local}
	\mathbf{v}^l_{i,j} = \mathbf{P}\left(\mathbb{AP}(\mathbf{f}_{i,j})\right), i=1,\cdots,N; j=1,\cdots,N_l
	\end{equation}
% 	We demonstrate that, guided by global information, even using the simple uniform horizontal part partition for local dictionary learning, we can achieve superior ReID performance to the state-of-the-arts.
 The global and local memory banks are then used to define global and local distance metrics respectively, which are further coupled with cross-camera encouragement term \citep{lin2020unsupervised} to define the total distance metric for pairwise similarity computation, as shown in Eq. (\ref{eq1}). According to the ranked similarity order, we identify the positives $\mathcal{K}_{+}$ and the negatives $\mathcal{K}_{-}$ that are used to define the contrastive loss. The details will be elaborated in the following subsection.

	\item \textit{Mixture memory bank} $\mathcal{M}^t$, which is required to store features of training images to define the contrastive loss, so as to maximize the similarity between representations of the anchor and the positives, while minimizing that of the anchor and the negatives. To improve the discrimination ability, we construct a mixture memory bank $\mathcal{M}^t$ which includes the fusion of the global and local features as keys. It is worth noting that, for the local feature in $\mathcal{M}^t$, instead of using the one defined in Eq. (\ref{local}), we turn to concatenate all $N_l$ ones to form a single local feature:
     \begin{equation}
     \label{conlocal}
	\mathbf{v}^l_i = \mathbb{CONCAT}\left(\{\mathbb{AP}(\mathbf{f}_{i,j})\}_{j=1}^{N_l}\right)
	\end{equation}
	
	The mixture memory bank $\mathcal{M}^t$ is initialized with all zeros. We update its keys corresponding to all positives by fusing with the global and local features of the anchor $\x_i$ progressively. Specifically, the positive keys are firstly updated with global feature $\mathbf{v}^g_i$:
		\begin{equation}
		\label{mixg}
	\mathcal{M}^t[k]=\|\frac{\mathcal{M}^t[k]+\mathbf{v}^g_i}{2}\|_2 , k\in \mathcal{K}_{+}
	\end{equation}
	which are further updated with local feature $\mathbf{v}^l_i$:
	\begin{equation}
	\label{mixl}
		\mathcal{M}^t[k]=\|\frac{\mathcal{M}^t[k]+\mathbf{v}^l_i}{2}\|_2 , k\in \mathcal{K}_{+}
	\end{equation}
	where $\|\cdot\|_2$ represents L2-normalization.
	In this way, the mixture memory bank jointly employs the global and local discriminative information.
	
% 	To form the mixture memory bank, a straightforward manner is to weighted combine the global and local features as its keys: $\mathcal{M}_i = w \mathbf{k}^g_i+(1-w)\mathbf{k}^l_i$, where $w$ is the weighting hyperparameter. In this work, we claim that weighted fusion would degrade the discriminative information, thus we turn to another manner that alternatively includes the global feature and the local feature as keys of $\mathcal{M}$ with the iterations.

	\end{itemize}
	
		\begin{algorithm}[htb]  
	\caption{ Network Training Flow.}  
	\label{alg:Framwork}  
	\begin{algorithmic}[1]  
		\Require  
		Training set $\mathcal{X} = \{\x_1, \x_2, ..., \x_{N}\}$;
		Global memory bank $\mathcal{M}^g$;
		Local memory bank $\mathcal{M}^l$;
		Mixture memory bank $\mathcal{M}^t$;
		network with parameter $\mathcal{F}(\boldsymbol{\theta};\x_i)$ ;
		initial learning rate $\boldsymbol{\gamma}$ ;
		\Ensure  
		The optimal parameters $\boldsymbol{\theta}^{*}$ 
		\For{$j$ = 1: num\_epochs} 
		\State Pick up training batch set $\{\x_i\}$.
		\For{$m$ = 1:$n$} 
		\State $\mathbf{f}^g_i = \mathbb{AP}(\mathbf{f}_i)$; $\mathbf{f}^l_{i,j} = \mathbb{AP}(\mathbf{f}_{i,j}), j=1,\cdots,N_l$
		\State $\mathbf{v}^g_i = \mathbf{P}\left(\mathbb{AP}(\mathbf{f}_i)\right)$
		\State $\mathbf{v}^l_{i,j} = \mathbf{P}\left(\mathbb{AP}(\mathbf{f}_{i,j})\right); j=1,\cdots,N_l$
	    \State $\mathbf{v}^l_i = \mathbb{CONCAT}\left(\{\mathbb{AP}(\mathbf{f}_{i,j})\}_{j=1}^{N_l}\right)$
		\If{$i<N_e$}
		\State $\mathcal{L}^g=\mathcal{L}^{init}(\mathbf{v}^g_i|\mathcal{M}^t)$
		\State $\mathcal{L}^l=\mathcal{L}^{init}(\mathbf{v}^l_i|\mathcal{M}^t)$
		\Else
		\State Calculate similarities between the anchor image and the other images.
		\For{$j$ = 1:$N$} 
		\If{$j\ne i$}
		\State  $\mathcal{S}^g(x_i,x_j)=\|\mathbf{v}^g_i - \mathcal{M}^g[j] \|_2$
		\State{$\mathcal{S}^l(x_i,x_j)=\frac{\sum_{k=1}^{N_l}\| \mathbf{v}^l_{i,k} - \mathcal{M}^l[j,k] \|_2}{N_l}$}
		\State $\mathcal{S}(\x_i,\x_j)=\beta \mathcal{S}^g(\x_i,\x_j) + (1-
\hspace{0.3cm}	\beta) \mathcal{S}^l(\x_i,\x_j) + \mathbb{CCE}(\x_i,\x_j)$
		\EndIf
		\EndFor
		\State Generate set with descending similarity $[\x_{j_1}, \x_{j_2},...,\x_{j_{N-1}}]$
		\State Sample positives $[\x_i, \x_{j_1}, \x_{j_2},..., \x_{j_{N_+}}]$
		\State Sample negatives $[\x_{j_{N_+}},...,\x_{j_{N_{-}+N_{+}}}]$
		\State $\mathcal{L}^g=\mathcal{L}(\mathbf{v}^g_i|\mathcal{M}^t)$
		\State $\mathcal{L}^l=\mathcal{L}(\mathbf{v}^l_i|\mathcal{M}^t)$
		\EndIf
		\State $\mathcal{L}^t(\boldsymbol{\theta})=(1-\lambda_p) \mathcal{L}^{g}+\lambda_p \mathcal{L}^{l}$
		\EndFor;
		\State$\boldsymbol{\theta}=\boldsymbol{\theta}-\boldsymbol{\gamma}*\mathbb{SGD}(\nabla_{\theta}{\mathcal{L}^t(\boldsymbol{\theta})})$;
		\State Update $M^g,M^l,M^t$
		\EndFor;
		
		\State $\boldsymbol{\theta^{*}}=\boldsymbol{\theta}$.
		
	\end{algorithmic}  
	
\end{algorithm}	
	
	\subsection{Positives and Negatives Sampling}
	
	As the name suggests, contrastive learning requires to obtain two opposing powers: for a given anchor sample, one power is to pull the anchor closer in representation space to other samples, which is known as the positive; while the other power is to push the anchor farther away from other samples, which is known as the negatives. To identify the positive and negative samples, we rely on the constructed global and local memory banks  $\mathcal{M}^g$ and $\mathcal{M}^l$ to compute pairwise similarity of samples, and apply two well-designed similarity metrics in the literature to this end. 
	
	Specifically, for an anchor image $\x_i$, we learn its global feature $\mathbf{v}^g_i$ and local feature $\{\mathbf{v}^l_{i,j}\}_{j=1}^{N_l}$ in the way as described in last subsection. The similarity between $\x_i$ and another image $\x_j$ are calculated by measuring Euclidean distance of feature vectors and keys of dual dictionary  $\mathcal{M}^g$ and $\mathcal{M}^l$. More precisely, we define the global and local distances as $\mathcal{S}^g$ and $\mathcal{S}^l$ respectively as follows: 
	\begin{equation}
		\mathcal{S}^g(\x_i,\x_j)=\|\mathbf{v}^g_i - \mathcal{M}^g[j] \|_2
	\end{equation}
	and
	\begin{equation}
		\mathcal{S}^l(\x_i,\x_j)=\frac{\sum_{k=1}^{N_l}\| \mathbf{v}^l_{i,k} - \mathcal{M}^l[j,k] \|_2}{N_l}
	\end{equation}

	Moreover, to encourage the consistency of images of the same person captured by different cameras, we add the cross-camera encouragement term (CCE) proposed in \citep{lin2020unsupervised} as a part of the similarity metric.  Set the camera IDs of person images $\x_i$ and $\x_j$ as $c_i$ and $c_j$ respectively, CCE is defined as
	\begin{equation}
		\mathbb{CCE}(\x_i,\x_j)=\left \{ \begin{array}{rcl}
		\lambda_c      &      {c_i=c_j}\\
		0     &     {c_i \neq c_j}\\
		\end{array} \right .
	\end{equation} 
	Finally, the total similarity metric $\mathcal{S}$ between $\x_i$ and $\x_j$ is formulated as:
	\begin{equation}
	\mathcal{S}(\x_i,\x_j)=\beta \mathcal{S}^g(\x_i,\x_j) + (1-
\beta) \mathcal{S}^l(\x_i,\x_j) + \mathbb{CCE}(\x_i,\x_j)
	\label{eq1}
	\end{equation}
	where $\beta$ is the trade-off parameter that balances the contribution of global and local similarity and is set 0.5.

	According to the defined total similarity metric, we perform positive and negative sampling. 	We rank the similarity order between the anchor and all training samples, according to which we divide the training set into three subsets: \textit{similar set}, 
\textit{borderline set}, and \textit{dissimilar set}. %We empirically set $N_{+} = 7$ and $N_{-} = 500$ respectively.
 Considering that each person may have multiple images in dataset, we choose to sample multiple positives for the anchor, as opposed to SimCLR \citep{chen2020simple} and MoCo \citep{he2020momentum} that use only a single positive to define the contrastive loss. Specifically, we consider samples in \textit{similar set} as the positives, whose index sets are denoted as $\mathcal{K}_{+} \in \mathbb{R}^{N_{+}}$. Moreover, we propose to select samples that are plausibly similar to the anchor as the negatives, so as to improve the discrimination ability of representation learning. Specifically, we consider in \textit{borderline set} as the negatives, whose index sets are denoted as $\mathcal{K}_{-}\in \mathbb{R}^{N_{-}}$.  This is different from the conventional contrastive learning strategies that take all samples except the positive as the negatives \citep{chen2020simple,he2020momentum}. And it is also different from an intuitive idea that chooses samples in \textit{dissimilar set} as the negatives. Since samples in \textit{dissimilar set} is already differentiable enough, learning on them cannot improve the discrimination ability of the model significantly.
Instead, we enforce the model to consider samples in \textit{borderline set}, which are hard cases with respect to  discrimination.	 We refer to as the proposed manner as selective contrastive learning.

\subsection{Contrastive Loss and Optimization }	

With the identified postives and negatives, we define the following contrastive loss function with respect to the mixture memory bank $\mathcal{M}^t$:
	\begin{equation}
	\label{loss}
	\mathcal{L}(\mathbf{v}_i|\mathcal{M}^t)=-\log \frac{\sum_{k \in \mathcal{K}_{+}}\exp(\mathbf{v}_i \cdot \mathcal{M}^t[k]/\tau)*\mu_k}{\sum_{k \in {\mathcal{K}_{+}}\cup\mathcal{K}_{-}}\exp(\mathbf{v}_i \cdot \mathcal{M}^t[k]/\tau)}
	\end{equation}
	where $\cdot$ represents dot product, $\tau$ is a temperature hyper-parameter \citep{2018Unsupervised}.  In order to emphsize the contribution of the most similar positive sample, we introduce the contribution factor $\mu_k$ for positive distance, which is defined as
		\begin{equation}
\mu_k=\left \{ \begin{array}{rcl}
\lambda_t    &      {k=i}\\
\frac{\alpha(1-\lambda_t)}{|\mathcal{K}_{+}|}   &     k \in \mathcal{K}_{+}\&k\neq i\\
\end{array} \right .
\end{equation} 
where $\alpha$ is the expanding coefficient and is set 1.75.

According to the loss form in Eq. (\ref{loss}), we calculate the global loss $\mathcal{L}^g$ and the local loss $\mathcal{L}^l$ with the global feature  $\mathbf{v}^g_i$ defined as Eq. (\ref{global}) and the local feature $\mathbf{v}^l_i$ defined as Eq. (\ref{conlocal}), respectively:
\begin{equation}
	\mathcal{L}^g=\sum_{i=0}^{N}\mathcal{L}(\mathbf{v}^g_i|\mathcal{M}^t)
\end{equation}
and
\begin{equation}
\mathcal{L}^l=\sum_{i=0}^{N}\mathcal{L}(\mathbf{v}^l_i|\mathcal{M}^t)
\end{equation}
And finally the total contrastive loss is defined as:
\begin{equation}
\label{totalloss}
	\mathcal{L}^t(\boldsymbol{\theta})=(1-\lambda_p) \mathcal{L}^{g}+\lambda_p \mathcal{L}^{l}
\end{equation}
where $\lambda_p$ is the trade-off parameter that balances the contributions of global and local losses and is set 0.5. In our framework, the parameters of $\mathbf{E}(\cdot)$ and $\mathbf{P}(\cdot)$ are collectively denoted as $\boldsymbol{\theta}$.

The optimal parameter $\boldsymbol{\theta}^{*}$ of ReID model $\mathcal{F}(\boldsymbol{\theta},\cdot)$ can be obtained by:
	\begin{equation}
	\boldsymbol{\theta}^{*} =\arg \min _{\boldsymbol{\theta}} \mathcal{L}^t(\boldsymbol{\theta})
	\end{equation}
	This minimization problem can be addressed by stochastic gradient descent (SGD):
	\begin{equation}
	\boldsymbol{\theta}=\boldsymbol{\theta}-\boldsymbol{\gamma}*\mathbb{SGD}(\nabla_{\theta}{\mathcal{L}^t(\boldsymbol{\theta})})
	\end{equation}
	where $\boldsymbol{\gamma}$ is the learning rate, and $\mathbb{SGD}(\nabla_{\theta}{\mathcal{L}^t(\boldsymbol{\theta})})$ represents the updated value based on SGD. The whole network training flow is summarized in Algorithm \ref{alg:Framwork}.

	It is worth noting that, at the beginning of network training, we set a few epochs to initialize the network and the memory banks. During this process, the positive is the anchor itself and the negatives are randomly chosen. The loss function in initialization stage is defined as follows:
\begin{equation}
\mathcal{L}^{\text{init}}=-\sum_{i=0}^{N}\log \frac{\exp(\mathbf{v}_i \cdot \mathcal{M}^t[i]/\tau)}{\sum_{k \in {\{i\}}\cup\mathcal{K}_{-}}\exp(\mathbf{v}_i \cdot \mathcal{M}^t[k]/\tau)} 
\end{equation}

% 	This contrastive loss serves as an unsupervised objective function for training the feature representation networks of the queries and keys.  By summing over all $i$, we define the final loss function as:
% 		\begin{equation}
% 	\mathcal{L} =  \sum_i\mathcal{L}(\mathbf{v}_i|\mathcal{D}_M)
% 	\end{equation}
% 	which is restricted from minimizing the denominator or maximizing the numerator without doing the other as well. As a result, we learn to map images of the same person to neighboring representations while mapping different ones to non-neighboring ones.

% 	\begin{figure*}[t] 
% 		\centering  

% 		\includegraphics[width=2.0\columnwidth]{./c1.png}

% 		\caption{Memory Bank Update Strategy}
% 		\label{c1}
% 	\end{figure*}

	\subsection{Memory Banks Update}
	
	Contrastive learning can be thought of as training an encoder for a dictionary look-up task \citep{he2020momentum}. In our method, as stated above, there are three dictionaries involved, including the global and local memory banks $\mathcal{M}^g$ and $\mathcal{M}^l$ that are jointly exploited to compute pairwise similarity, and the mixture memory bank $\mathcal{M}^t$ that is used to define constrative loss. To facilitate contrastive unsupervised learning, these three dictionaries should be dynamic, \textit{i.e.}, be updated on-the-fly to provide evolutionary keys during training.
	
	In this work, we propose to use different update strategies for the global and local memory banks and the mixture memory bank, considering that they serve for different purposes. Specifically, for $\mathcal{M}^g$ and $\mathcal{M}^l$, we only update the key corresponding to the anchor $\x_i$ by fusing with the newest global and local feature of $\x_i$ respectively:
	\begin{equation}
		\mathcal{M}^g[i]=\|\frac{\mathcal{M}^g[i]+\mathbf{v}^g_i}{2}\|_2,
	\end{equation}
		and
	\begin{equation}
		\mathcal{M}^l[i]=\|\frac{\mathcal{M}^l[i]+\mathbf{v}^l_i}{2}\|_2,
	\end{equation}

	For the mixture memory bank, we will update the key corresponding to the anchor $\x_i$ and its positives. The update strategy  is consistent with the construction strategy, as formulated in Eq. (\ref{mixg}) and Eq. (\ref{mixl}).

% 	\subsection{Test for ReID}
	
% 	For completeness, in this subsection, we interpret how our scheme works in the testing stage.
	
% 	According to Eq. (\ref{global}) and Eq. (\ref{conlocal}), the learned representation model is applied on the query image $\x^q$ and the gallery set $\hat{\mathcal{X}} = \{\hat{\x}_1, \hat{\x}_2, ..., \hat{\x}_{M}\}$ to extract their global and local features. We then compute the total similarity metric defined in Eq. (\ref{eq1}) for pairs of sample $(\x^q, \hat{\x}_i), i = 1,\cdots,M$. By ranking the distance between features of the query and all gallery images, we derive the re-identified person images. 
	
	\subsection{Why Using Three Dynamic Dictionaries?}
	In our framework, three dynamic dictionaries are used, including the global, local and mixture memory banks. We discuss here about the necessity of using these three dictionaries.
	
 	In pairwise similarity computation, we use both the global and local memory banks instead of the mixture one. This is because the keys of the mixture memory bank are the fusion of the global and local features. The fusion operation would remove some useful information. In order to preserve fine information, we thus use both the original global and local features, which construct the global and local memory banks. 
%are built for . In this process, the ``accurate" similarity is critical for identifying the positives and the negatives. Thus we are supposed to keep the maximum instance information for similarity computation. In such case, the separate global and local memory banks instead of mixture memory bank and only updating the key corresponding to the anchor $\x_i$  could provide more precise metrics for ranking the similarity order.
	
In defining the contrastive loss, we use the mixture memory bank instead of the global and local ones. For the task of person ReID,  we except the model to generate similar feature representation for the samples with same identity and dissimilar representation otherwise. According to this principle, the mixture memory bank is tailored, in which we update the keys corresponding to the positives of the anchor $\x_i$. This would encourage the anchor and its positives have similar representation and update the keys with global features and local features. It also could provide more discriminative ability to pull the similar samples closer and push dissimilar samples apart away.

	\section{Experiments}

			\begin{figure*}[t] 
		\centering  

		\subfigtopskip=2pt 
		\subfigbottomskip=2pt 
		\subfigcapskip=-5pt 
		\subfigure[]{
			\label{lc}
			\includegraphics[width=0.3\linewidth]{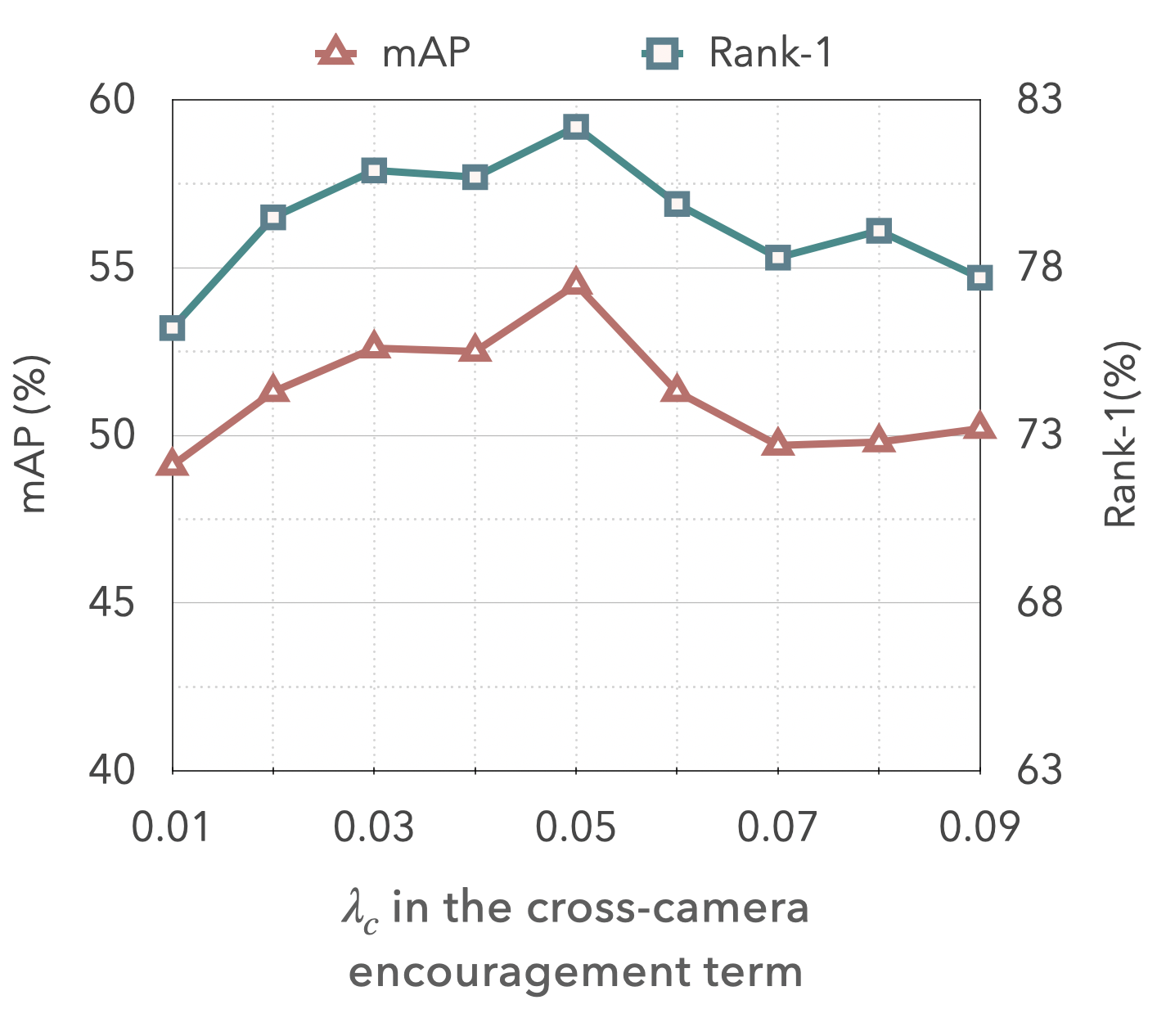}}
		\subfigure[]{
			\label{lt}
			\includegraphics[width=0.3\linewidth]{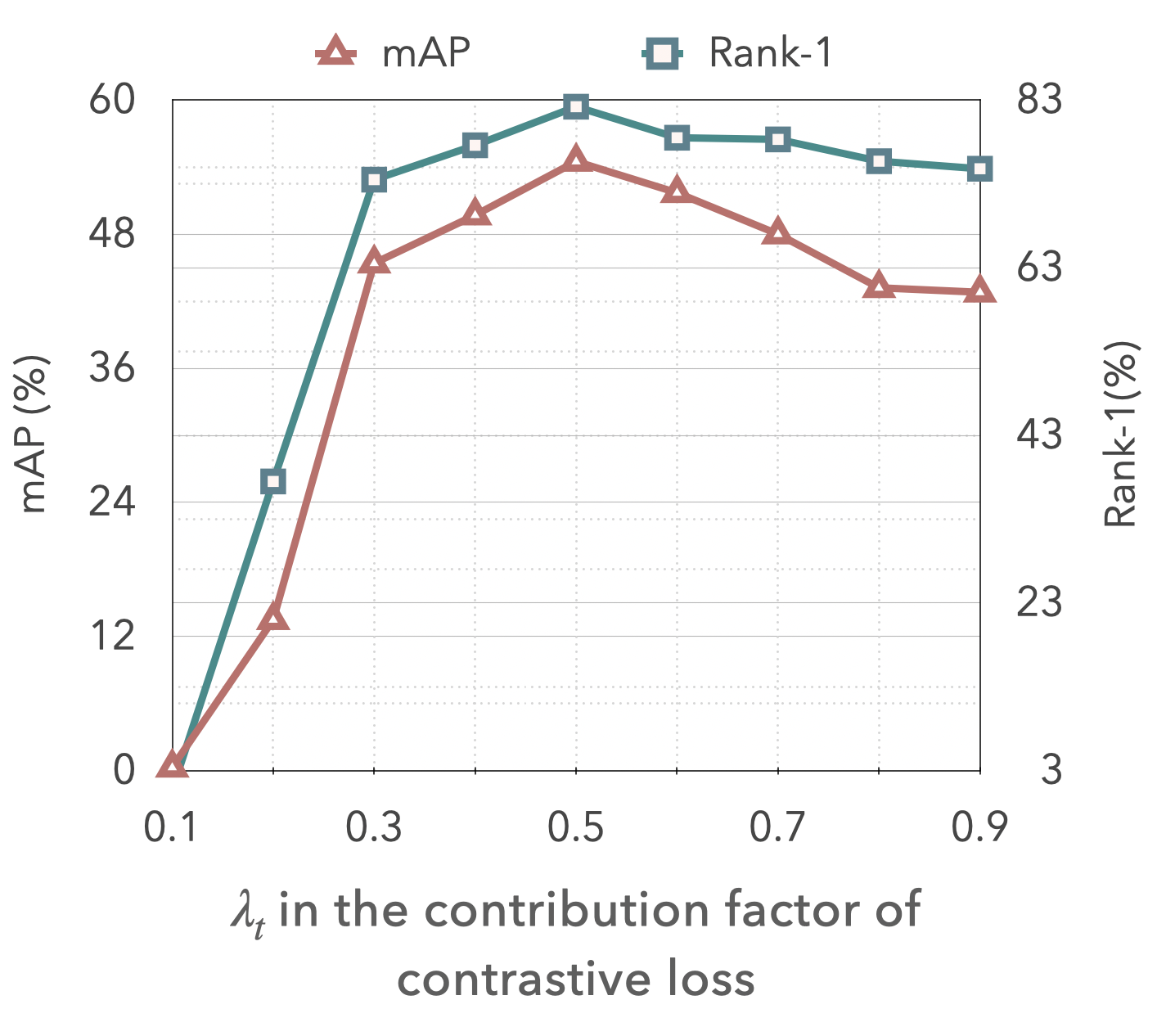}}

		\subfigure[]{
			\label{n+}
			\includegraphics[width=0.3\linewidth]{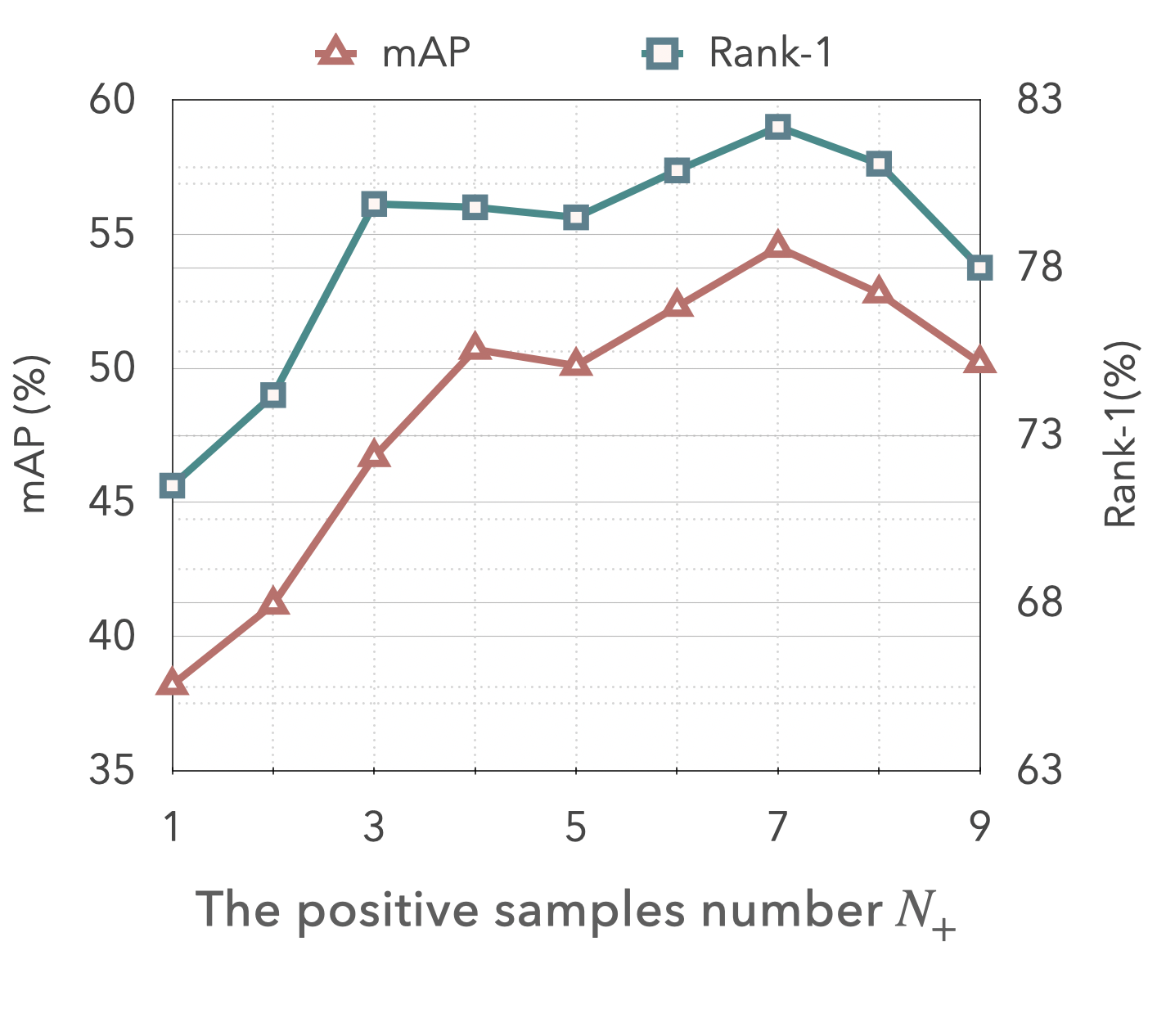}}	
		\subfigure[]{
			\label{n-}
			\includegraphics[width=0.3\linewidth]{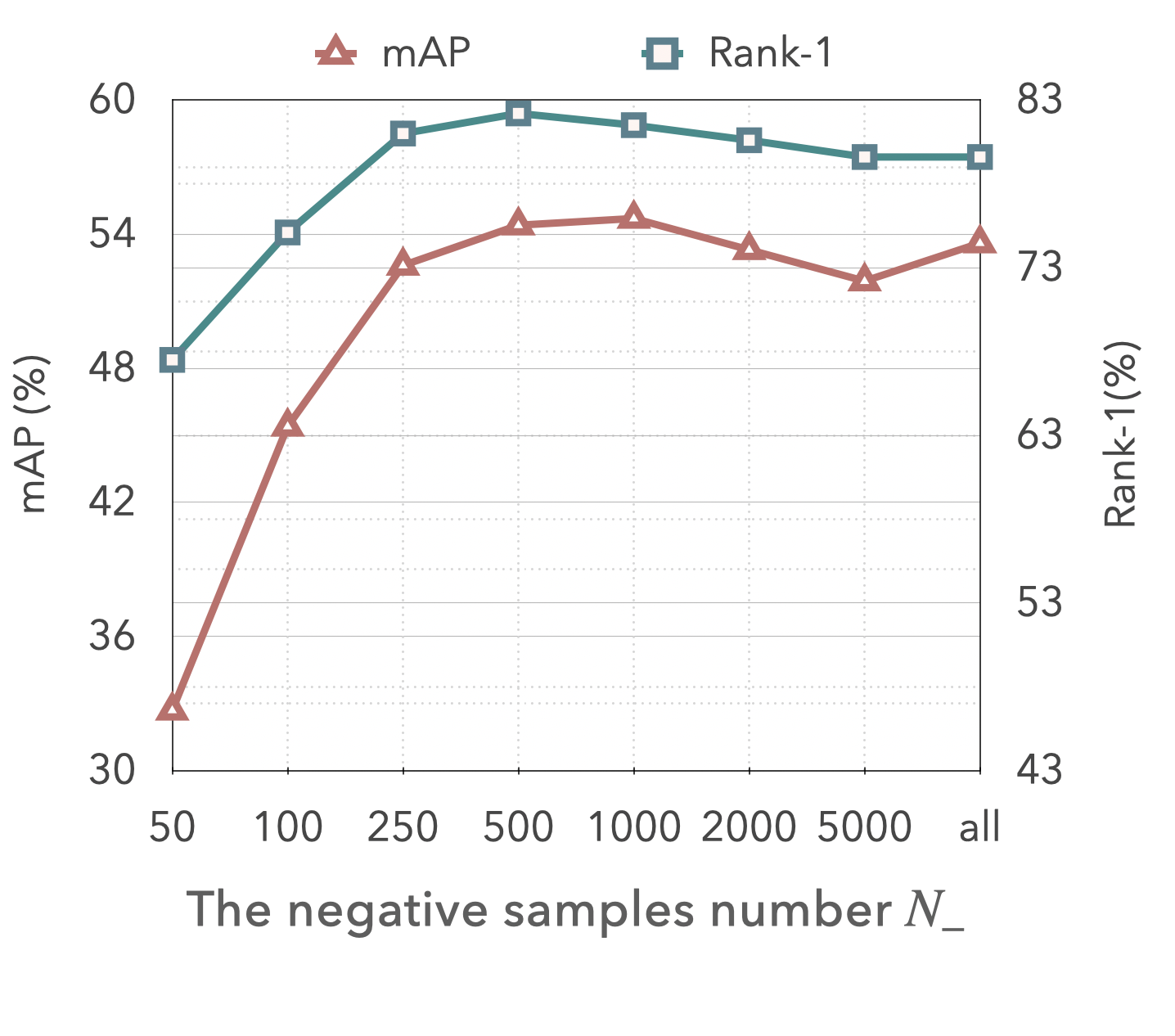}}

		\caption{Experimental analysis about hyper-parameters setting}
		\label{aklt}

	\end{figure*}
	
	In this section, we provide extensive experimental results to demonstrate the superior performance of our method. 

	\subsection{Dataset}
	
	We evaluate our method on five widely used image and video person ReID datasets, including: 
	\begin{itemize}
		\item \textit{Market1501} \citep{market}, which consists of 32,668 images of 1,501 identities under 6 cameras; 
       \item \textit{DukeMTMC-ReID} \citep{ristani2016MTMC}, which contains 1,812 identities and 36,411 images under 8 cameras;
       \item \textit{MSMT17} \citep{2018Person}, which contains 126,411 person images of 4101 identities under 15 cameras. It is a more challenging dataset due to the effect of substantial variations of scene and lighting. 
       \item \textit{DukeMTMC-VideoReID} \citep{videoreid}, which is a video-based ReID dataset containing 2,196 tracklets of 702 identities for training, 2,636 tracklets of other 702 identities for testing;
       \item \textit{MARS} \citep{mars}, which is a video-based dataset for person ReID containing 17,503 video tracklets of 1,261 identities. 
       \end{itemize}
       The first three datasets are image-based, and the last two ones are video-based. We test both image and video based to comprehensively evaluate the performance of our method.

	\subsection{Experimental Setting}
% We follow the same experiment settings with \citep{lin2020unsupervised,lin2019bottom}. All the experiments are based on PyTorch. The number of the local parts $N_l$ and the embedding feature vectors size are set 8 and 2048. The global memory bank and the local memory bank are initialized with the embedding feature vectors generated by the network. And the mixture memory bank is initialized to all zeros. Then number of initial epochs  $N_e$ is  set 3.
	
We follow the same experimental setting as \citep{lin2020unsupervised}. All experiments are implemented on PyTorch.	The input images are resized to 256*128 and we use random horizontal flip as the data argument strategy. We adopt SGD with momentum as 0.9 to optimize the model. The learning rate is set as 1e-3. The training epoch for image-based dataset is set as 50 and for the video-based dataset is set as 60. The batch size is set as 8. For the video-based dataset, we randomly sample four frames during training, and all frames during testing, in the tracklet. We take the average feature of all frames within a tracklet to be the tracklet feature.
	
In the proposed framework, there are a few hyper-parameters involved. We set $\lambda_c = 0.005$ in the cross-camera encouragement term, $\beta = 0.5$ in the total similarity metric, $\lambda_t = 0.5$ and $\alpha = 1.75$ in the contribution factor of contrastive loss, the positive samples number $N_+ = 7$ and the negative samples number $N_- = 500$, and  $\lambda_p = 0.5$ in the total contrastive loss. The experimental analysis about hyper-parameters setting can be found in Fig. \ref{aklt}, which is conducted on Market-1501.

% 	We set $\beta=0.5,\alpha=1.75$ to calculate total similarity distance and contrastive loss respectively. $\lambda_c$ and $\lambda_g$ are set as 0.05 and 0.5,  respectively. All experiments are implemented on PyTorch. Extensive parameters analysis will be presented in the following.
	
	\subsection{Comparison with the state-of-the-arts}

				\begin{table*}[htbp]
			\resizebox{2.1\columnwidth}{!}{
\begin{tabular}{c||c||c||c|c|c|c|c||c|c|c|c|c}
\hline
\multirow{2}{*}{\textbf{Method}} & \multirow{2}{*}{\textbf{Reference}} & \multirow{2}{*}{\textbf{Setting}} & \multicolumn{5}{c||}{\textbf{Market-1501}}                              & \multicolumn{5}{c}{\textbf{DukeMTMC-reID}}                            \\ \cline{4-13} 
                                 &                                     &                                   & Source & Rank-1        & Rank-5        & Rank-10       & mAP           & Source & Rank-1        & Rank-5        & Rank-10       & mAP           \\ \hline\hline
EUG     \citep{2018Exploit}                         & CVPR'2018                           & OneEx                             & Market & 49.8          & 66.4          & 72.7          & 22.5          & Duke   & 45.2          & 59.2          & 63.4          & 24.5          \\ \hline
ATNet    \citep{liu2019adaptive}                        & CVPR'2019                           & UDA                               & Duke   & 55.7          & 73.2          & 74.9          & 25.6          & Market & 45.1          & 59.5          & 64.2          & 24.9          \\ \hline
ProLearn    \citep{2019Progressive}                     & TIP'2019                            & OneEx                             & Market & 55.8          & 72.3          & 78.4          & 26.2          & Duke   & 48.8          & 63.4          & 68.4          & 28.5          \\ \hline
SPGAN  \citep{deng2018image}                          & CVPR'2018                           & UDA                               & Duke   & 58.1          & 76.0          & 82.7          & 26.7          & Market & 46.9          & 62.6          & 68.5          & 26.4          \\ \hline
TJ-AIDL    \citep{wang2018transferable}                      & CVPR'2018                           & UDA                               & Duke   & 58.2          & -             & -             & 26.5          & Market & 44.3          & -             & -             & 23.0          \\ \hline
BUC    \citep{lin2019bottom}                          & AAAI'2019                           & Unsup                             & None   & 61.0          & 71.6          & 76.4          & 30.6          & None   & 40.2          & 52.7          & 57.4          & 21.9          \\ \hline
HHL    \citep{zhong2018generalizing}                          & ECCV'2018                           & UDA                               & Duke   & 62.2          & 78.8          & 84.0          & 31.4          & Market & 46.9          & 61.0          & 66.7          & 27.2          \\ \hline
DBC     \citep{ding2019dispersion}                         & BMVC'2019                           & Unsup                             & None   & 69.2          & 83.0          & {\ul 87.8}    & 41.3          & None   & 51.5          & {\ul 64.6}    & {\ul 70.1}    & 30.0          \\ \hline
SNR    \citep{2020Style}                 & CVPR'2020                           & UDA                               & Duke   & 66.7          & -             & -             & 33.9          & Market & 55.1          & -             & -             & 33.6          \\ \hline
SSLR           \citep{lin2020unsupervised}                  & CVPR'2020                           & Unsup                             & None   & {\ul 71.7}    & {\ul 83.8}    & 87.4          & 37.8          & None   & 52.5          & 63.5          & 68.9          & 28.6          \\ \hline
MMCL       \citep{wang2020unsupervised}                 & CVPR'2020                           & Unsup                             & None   & 66.6          & -             & -             & 35.3          & None   & 58.0          & -             & -             & 36.3          \\ \hline
TSSL      \citep{2020Tracklet}                       & AAAI'2020                           & Unsup                             & None   & 71.2          & -             & -             & {\ul 43.3}    & None   & {\ul 62.2}    & -             & -             & {\ul 38.5}    \\ \hline
Ours                             & This paper                          & Unsup                             & None   & \textbf{82.2} & \textbf{89.9} & \textbf{92.6} & \textbf{54.4} & None   & \textbf{69.9} & \textbf{79.7} & \textbf{82.2} & \textbf{47.2} \\ \hline
\end{tabular}
}
		\caption{The  evaluation results with respect to rank-k/mAP on image-based dataset Market-1501 and DukeMTMC. The best and the second ones are highlighted by bold and underline.}
		\label{mardu}
	\end{table*}
	
	Our method is comprehensively compared with state-of-the-art unsupervised domain adaption based (UDA), one example based (OneEx) and unsupervised learning based (Unsup) methods. The comparison is conducted on both image-based and video-based datasets.
	%Two imaged based datasets: Market-1501 \citep{zheng2015scalable} and DukeMTMC-reID \citep{ristani2016performance} and two video-based datasets:  MARS \citep{zheng2016mars} and DukeMTMC-VideoReID \citep{wu2018exploit} are used for evaluation. The numerical results are from their own paper. (MMCL+MPLP's results are taken from their ablation study, where  the CamStyle \citep{zhong2018camera} for data gmentation is unused, for fair comparison. )  
	
	\begin{itemize}
		\item\textbf{Evaluation on Image-based Datasets:}
	The comparisons with the state-of-the-art algorithms are conducted on Market-1501, DukeMTMC-ReID and MSMT17, as shown in Table \ref{mardu}. 
% 	The comparison study group includes:
% 	\begin{itemize}
% 		\item Unsupervised domain adaption (UDA) based: 1$)$ Adaptive transfer network, ATNet \citep{liu2019adaptive}; 2$)$ Preserved self-similarity and domain-dissimilarity, SPGAN \citep{deng2018image}; 3$)$ Transferable joint attribute-identity deep learning, TJ-AIDL \citep{wang2018transferable}; 4$)$ Hetero-and homogeneously model, HHL \citep{zhong2018generalizing}
% 		\item One example (OneEx) based: 1$)$ Exploit the unknown gradually, EUG \citep{2018Exploit}; 2$)$ Progressive learning, ProLearn \citep{2019Progressive}
% 		\item Unsupervised learning (Unsup) based: 1$)$ Local maximal occurrence representation and metric learning, LOMO \citep{liao2015person}; 2$)$ Scalable person re-identification, BOW \citep{zheng2015scalable}; 3$)$ Joint detection and identification feature learning, OIM \citep{xiao2017joint}; 4$)$ A bottom-up clustering approach, BUC \citep{lin2019bottom} 5$)$ Dispersion based clustering, DBC \citep{ding2019dispersion} 6$)$ Softened Similarity Learning, SSLR \citep{lin2020unsupervised} 7$)$ Multi-label Classification, MMCL \citep{wang2020unsupervised}
% 		\end{itemize}
It can be found that, under the same setting, our method achieves the best accuracy on both Market-1501 and DukeMTMC-ReID among the 14 compared methods with respect to four performance evaluation metrics: rank-1, rank-5, rank-10 and mAP. 

On Market-1501,  we obtain the best performance among the compared methods with rank-1 = 82.2\% and mAP = 54.4\%.
SNR \citep{2020Style}, SSLR \citep{lin2020unsupervised}, MMCL \citep{wang2020unsupervised} and TSSL \citep{2020Tracklet} are four latest methods on unsupervised person ReID.  Compared with SNR, we achieve 15.5\% and 20.5\% improvement on rank-1 accuracy and mAP; compared with SSLR, the gains are 10.5\% and 16.6\% respectively; compared with MMCL, the gains are 15.6\% and 19.1\% respectively; compared with TSSL, the gains are 11\% and 11.1\% respectively.
On DukeMTMC-ReID, our method also works the best and achieves accuracy improvement by a large margin. Compared with the second best performed method, we achieve 7.7\% and 8.7\% improvement on rank-1 accuracy and mAP respectively.
On MSMT17, as shown in Table \ref{msmt}, compared with the best UDA method SSG, we achieve 9.2\% improvement on rank-1. Compared with fully unsupervised method MMCL, we achieve 6.0\% and 2.1\% improvement on rank-1 and mAP respectively. 

The impressive performance demonstrates that the proposed selective contrastive learning framework is able to learn a powerful discrimination model.

% 	The experiment results are illustrated in Table \ref{mardu}. We achieve the highest results with $79.6\%$ in Rank-1 and $53.6\%$ in mAP on Market-1501 and  $65.8\%$ in Rank-1 and $42.4\%$ in mAP on DukeMTMC-reID. Compared with best OneEx based method ProLearn, we have $23.8\%$ and $27.4\%$ improvement  in Rank-1 and  mAP  on Market-1501 and $17.0\%$ and $13.9\%$ improvement  in Rank-1 and  mAP  on DukeMTMC. Compared with best UDA based method HHL, we gains over $17.4\%$ and $22.2\%$ points in Rank-1 and  mAP  on Market-1501 and $18.9\%$ and $15.2\%$  points in Rank-1 and  mAP  on DukeMTMC. We improved $7.9\%$/$15.8\%$  and  $13.0\%$/$18.3\%$ points in  Rank-1/mAP  compared with SSLR and MMCL+MPLP on Market-1501 and $13.3\%$/$13.8\%$  and  $7.8\%$/$6.1\%$ points in  Rank-1/mAP  compared with SSLR and MMCL+MPLP on DukeMTMC. These prove the suprior of our framework.
	
\item \textbf{Evaluation on Video-based Datasets:}
We further compare our method with the state-of-the-art algorithms on the two video-based datasets: DukeMTMC-VideoReID and MARS. The comparison results are shown in Table \ref{dukevideo_mars}.	On DukeMTMC-VideoReID, we obtain rank-1 = 82.3\%, mAP = 78.4\%, which are best among all methods. Compared with SSLR, the gains are 5.8\% and 9.1\% respectively; compared with TTSL, the gains are 8.3\% and 13.8\% respectively. On MARS,  we obtain rank-1 = 66.7\%, mAP = 46.8\%, which are also the best results.

\begin{table*}[htbp]
\centering			\resizebox{1.45\columnwidth}{!}{
				\begin{tabular}{c||c||c||c|c|c|c|c}
					\hline
					\multirow{2}{*}{\textbf{Method}} & \multirow{2}{*}{\textbf{Reference}} & \multirow{2}{*}{\textbf{Setting}} & \multicolumn{5}{c}{\textbf{MSMT17}}                                                   \\ \cline{4-8} 
					&                                     &                                   & Source & Rank-1        & Rank-5        & Rank-10       & mAP                    \\ \hline\hline
					PTGAN  \cite{2018Person}                           & CVPR'2018                          & UDA                             & Market   & 10.2 & - & 24.4 & 2.9 \\ \hline
					ECN    \cite{2020Invariance}                        & CVPR'2020                          & UDA                             & Market   & 25.3 & 36.3 & 42.1 & 8.5 \\ \hline
					SSG     \cite{2018Self}                       & ICCV'2019                          & UDA                             & Market   & 31.6 & - & 49.6 & 13.2 \\ \hline\hline
					PTGAN    \cite{2018Person}                         & CVPR'2018                          & UDA                             & Duke   & 11.8 & - & 27.4 & 3.3 \\ \hline
					ECN    \cite{2020Invariance}                        & CVPR'2020                          & UDA                             & Duke   & 30.2 & 41.5 & 46.8 & 10.2 \\ \hline
					SSG   \cite{2018Self}                         & ICCV'2019                          & UDA                             & Duke   & 32.2 & - & {\ul 51.2} & {\ul 13.3} \\ \hline\hline
					MMCL       \cite{2020Unsupervised}                      & CVPR'2020                          & Unsup                             & None   & {\ul 35.4} & {\ul 44.8} & 49.8 & 11.2  \\ \hline
					Ours                             & This paper                          & Unsup                             & None   & \textbf{41.4} & \textbf{53.6} & \textbf{58.7}  & \textbf{13.3} \\ \hline
				\end{tabular}
			}
			\caption{The  evaluation results with respect to rank-k/mAP on image-based dataset MSMT17. The best and the second ones are highlighted by bold and underline.}
			\label{msmt}
		\end{table*}

		\begin{table*}[htbp]
\resizebox{2.0\columnwidth}{!}{
\begin{tabular}{c||c||c||c|c|c|c||c|c|c|c}
\hline
\multirow{2}{*}{\textbf{Method}} & \multirow{2}{*}{\textbf{Reference}} & \multirow{2}{*}{\textbf{Setting}} & \multicolumn{4}{c||}{\textbf{DukeMTMC-VideoReID}}              & \multicolumn{4}{c}{\textbf{MARS}}                            \\ \cline{4-11} 
                                 &                                     &                                   & Rank-1        & Rank-5        & Rank-10       & mAP           & Rank-1        & Rank-5        & Rank-10       & mAP           \\ \hline
RACE     \citep{ye2018robust}                        & ECCV'2018                           & OneEx                             & -             & -             & -             & -             & 43.2          & 57.1          & 62.1          & 24.5          \\ \hline
DAL     \citep{chen2018deep}                         & BMVC'2018                           & Unsup                             & -             & -             & -             & -             & 49.3          & 65.9          & 72.2          & 23.0          \\ \hline
BUC    \citep{lin2019bottom}                          & AAAI'2019                           & Unsup                             & 76.2          & 88.3          & 91.0          & 68.3          & 57.9          & 72.3          & 75.9          & 34.7          \\ \hline
EUG      \citep{wu2018exploit}                        & CVPR'2018                           & Unsup                             & 72.7          & 84.1          & -             & 63.2          & 62.6          & 74.9          & -             & 42.4          \\ \hline
SSLR  \citep{lin2020unsupervised}                           & CVPR'2020                           & Unsup                             & {\ul 76.4}    & {\ul 88.7}    & {\ul 91.0}    & {\ul 69.3}    & {\ul 62.8}    & \textbf{77.2} & \textbf{80.1} & {\ul 43.6}    \\ \hline
TTSL  \citep{2020Tracklet}                           & AAAI'2020                           & Unsup                             & 73.9          & -             & -             & 64.6          & 56.3          & -             & -             & 30.5          \\ \hline
Ours                             & This paper                          & Unsup                             & \textbf{82.2} & \textbf{93.2} & \textbf{95.2} & \textbf{78.4} & \textbf{66.6} & {\ul 77.0}    & {\ul 79.8}    & \textbf{46.6} \\ \hline
\end{tabular}
}
		
	\caption{The  evaluation results with respect to rank-k/mAP on video-based dataset DukeMTMC-VideoReID and MARS. The best and the second ones are highlighted by bold and underline.}
		\label{dukevideo_mars}
\end{table*}

% 	For video based dataset, the comparison study group includes:
% 		\begin{itemize}
% 		\item OneEx-based: 1$)$ Dynamic label graph matching, DGM+IDE \citep{ye2017dynamic} 2$)$  Stepwise metric promotion, Stepwise \citep{liu2017stepwise} 3$)$ Robust anchor embedding, RACE \citep{ye2018robust}
% 		4$)$ Exploit the unknown gradually, EUG \citep{2018Exploit};
% 		\item Unsupervised: 1$)$ Joint detection and identification feature learning, OIM \citep{xiao2017joint};  2$)$  Deep association learning, DAL\citep{chen2018deep} 3$)$ A bottom-up clustering approach, BUC \citep{lin2019bottom}  4$)$ Softened Similarity Learning, SSLR \citep{lin2020unsupervised} 
% 		\end{itemize}
	
% 	The results are shown in Table \ref{dukevideo} and Table \ref{mars}. We will discuss our results with MARS in the next section. 
% 	We achieve $80.2\%$/ $77.2\%$ in Rank-1/mAP and gains over $24.0\%$/ $21.4\%$ points compared with OneEx based methods. Alough Lin \textit{et al.}  \citep{lin2020unsupervised} found BUC have quite high results with DukeMTMC-VideoReID under the unsupervised settings and their results are close to BUC. We still have $3.8\%$/$7.9\%$  improvement in Rank-1/mAP compared with their method.	
\end{itemize}

	\begin{table}[htbp]
	\centering
	\resizebox{0.9\columnwidth}{!}{
\begin{tabular}{c||c|c||c|c}
\hline
\multirow{2}{*}{\textbf{Scenarios}} & \multicolumn{2}{c||}{\textbf{Market-1501}} & \multicolumn{2}{c}{\textbf{DukeMTMC}} \\ \cline{2-5} 
                                 & mAP                & Rank-1               & mAP               & Rank-1             \\ \hline
Global feature only                      & 38.3               & 62.0                 & 26.9              & 41.2               \\ \hline
Local feature only                       & 42.4               & 70.8                 & 34.1              & 57.8               \\ \hline
Our joint usage                             & \textbf{54.4}               & \textbf{82.2}                 & \textbf{47.2}              & \textbf{69.9}               \\ \hline
\end{tabular}
	}
\caption{The ablation study about the influence of local and global features to the final performance}
\label{ablationm}
\end{table}
	
% 	\begin{table}[htbp]
% 	\centering
% 	\resizebox{0.6\columnwidth}{!}{
% \begin{tabular}{|c|c|c|c|c|}
% \hline
% \multirow{2}{*}{\textbf{Method}} & \multicolumn{2}{c|}{\textbf{Market-1501}} & \multicolumn{2}{c|}{\textbf{DukeMTMC}} \\ \cline{2-5} 
%                                  & mAP                & Rank-1               & mAP               & Rank-1             \\ \hline
% Global only                      & 18.3               & 42.0                 & 16.9              & 31.2               \\ \hline
% Local only                       & 42.4               & 70.8                 & 34.1              & 57.8               \\ \hline
% Without Mix                      & 20.6               & 43.0                 & 5.7               & 12.1               \\ \hline
% Ours                             & 54.4               & 82.2                 & 47.2              & 69.9               \\ \hline
% \end{tabular}
% 	}
% \caption{The ablation study about influence with mixture representation}
% \label{ablationm}
% \end{table}

	\begin{table}[htbp]
	\centering
	\resizebox{1.0\columnwidth}{!}{
	\begin{tabular}{c||c|c||c|c}
\hline
\multirow{2}{*}{\textbf{Scenarios}} & \multicolumn{2}{c||}{\textbf{Market-1501}} & \multicolumn{2}{c}{\textbf{DukeMTMC}} \\ \cline{2-5} 
                                 & mAP                & Rank-1               & mAP               & Rank-1             \\ \hline
$N_+=1,N_-=all$                      & 37.1               & 70.9                 & 37.9              & 60.3               \\ \hline
$N_+=7,N_-=all$                        & 53.6               & 79.6                & 42.4             & 65.8               \\ \hline
Ours ($N_+=7,N_-=500$)                              & \textbf{54.4}               & \textbf{82.2}                 & \textbf{47.2}              & \textbf{69.9}               \\ \hline
\end{tabular}
	}
\caption{The ablation study about the influence of positives and negatives to the final performance}
\label{ablationc}
\end{table}

	\subsection{Ablation Study}
In this section, we provide ablation study about the two main contributions of this work: selective contrastive learning and joint usage of global and local features. The experiments are conducted on Market-1501 and DukeMTMC-ReID. The results are reported in Table \ref{ablationm} and Table \ref{ablationc}.
\begin{itemize}
    \item 
	\textbf{Influence of local and global features to the final performance:} 
	We first study the role of global and local features to the final performance. We investigate three scenarios: 
	\begin{itemize}
    \item \textbf{Global feature only}: in this case, we use global feature only in pairwise similarity computation and contrastive loss. This is achieved by setting $\beta=1$ and $\lambda_p=0$ in Eq. (\ref{eq1}) and Eq. (\ref{totalloss}), respectively. The mixture memory bank used in contrastive loss stores global features only. 
    \item \textbf{Local feature only}: in this case, we use local feature for these purposes. This is achieved by setting $\beta=0$ and $\lambda_p=1$. The mixture memory bank stores local features only. 
    \item \textbf{Our joint usage}. This is what we do in this work, \textit{i.e.}, jointly using global and local features.
    \end{itemize}
	
	From Table \ref{ablationm}, it can be found that, the joint usage strategy shows the best performance, which achieves improvements with respect to mAP and rank-1 accuracy with a large margin compared with strategies with global or local feature only. This demonstrates that our proposal of jointly using global and local features in dictionary construction is reasonable and works well.

	\item
	\textbf{Influence of positives and negatives to the final performance:}
	In our proposed selective contrastive learning,  we leverage multiple positives and selective negatives in contrastive loss definition. Here we study the role of multiple positives and selected negatives to the final performance, and investigate the following three scenarios: \begin{itemize}
	    \item 
	 \textbf{$N_+=1,N_-=all$}: %Remember $N_+$ represents the number of positives, and $N_-$ represents the number of negatives.
	This case means that, a single positive is used, and all samples except the positive are used as the negatives, which is what MoCo does \citep{he2020momentum}. 
	\item \textbf{$N_+=7,N_-=all$}: This case means that multiple positives are used and all samples except the positives are used as the negatives. 
	\item \textbf{$N_+=7,N_-=500$}: This is what our method does. We use 7 similar samples as positives, and 500 borderline similar samples as negatives. 
	\end{itemize}
	
	From Table \ref{ablationc}, it can be found that, when $N_-=all$, using $N_+=7$ positives can significantly improve rank-1 and mAP accuracy compared with using a single positive. This result demonstrates that our proposal of using multiple positives in contrastive learning is useful. Moreover, when $N_+=7$, using selected $N_-=500$ negatives achieves higher  rank-1 and mAP accuracy than that using $N_-=all$ negatives. This result demonstrates that our proposed selective choice strategy of negatives is also helpful for contrastive learning.
% 	We test our framework with global features by setting $\lambda_p=0.0, \beta=1.0$ and the memory bank used for calculating loss will store and be updated by global features only. Compared with the global features only, our method gains over $37.6\%$/$35.3\%$ in Rank-1/mAP on Market-1501 and  $34.6\%$/$25.5\%$ in Rank-1/mAP on DukeMTMC.
	
% 	 \item 
% 	\textbf{Local only:} We test our  framework with local features by setting $\lambda_p=1.0, \beta=0.0$ and the memory bank used for calculating loss will store and be updated by local features only. And our method will have $8.8\%$/$11.2\%$ in Rank-1/mAP on Market-1501 and  $8.0\%$/$8.3\%$ in Rank-1/mAP on DukeMTMC. The local feature works much better than global feature, but it still hardly achieve more better results due to the indepdent local parts.  
	
% 	 \item 
% 	\textbf{Without Mix:} For experiment setting with mixture memory bank removed, relative parameter will be maintained. The global loss $\mathcal{L}^g$ will use the memory bank which stores global embedding feature vectors and The local loss $\mathcal{L}^l$ will use the memory bank which stores concatenate local embedding feature vectors. Under such settings, there will be no connection and might cause extremely inconsistent between the global features and the local features. As shown in the tabele, if the global features and the local features are independently trained, the results will be lower than the local features only on Market-1501 and will be inefficient on DukeMTMC.

\end{itemize}

% \subsection{Failure case}
% As shown in Table \ref{mars}, on this dataset, our framework is not obviously suprior than other methods compared with other datasets. There exits many pictures in MARS, as shown in Fig \ref{mf}, the background has taken too much space in the pictures. Accroding to our approach, when we attemp to adapt local feature, it will bring in indenty-irrelevant information, causing degradation in performance. But such condition could be avoided by a fine detection box, where the person could be well parted from the background.

% 	\begin{figure}[t] 
	
% 	\includegraphics[width=0.9\linewidth]{./mf1.png}
% 	\caption{Conflict pictures in the MARS dataset.}
% 	\label{mf}\end{figure}

	\section{Conclusion}
	In this work, we presented a novel unsupervised person ReID scheme based on selective contrastive learning. We propose to use multiple positives and adaptively selected negatives for defining the contrastive loss,  so as  to learn a  feature  embedding  model  with stronger discriminative representation ability. We define three dynamic dictionaries for pairwise similarity computation and contrastive loss definition, which jointly leverage the global and local discriminative information. Experimental results show that our proposed method outperforms the state-of-the-art algorithms.
	\ifCLASSOPTIONcaptionsoff
  \newpage
\fi
	
% 		\section{Acknowledgements}
% 	This work is supported by the National Science Foundation of China under Grants 61922027, 61971165 and  61932022.
{
\bibliographystyle{IEEEtran}
\bibliography{test}
}

\end{document}